\pdfoutput=1

\documentclass[11pt]{article}

\usepackage[]{acl}

\usepackage{times}
\usepackage{latexsym}
\usepackage{kotex}
\usepackage{graphicx}
\usepackage{amsmath}
\usepackage{amssymb}
\usepackage{booktabs}
\usepackage{adjustbox}
\usepackage{geometry}
\usepackage{algorithm}
\usepackage{algorithmic}
\usepackage{tabularx}
\usepackage{longtable}
\usepackage[T1]{fontenc}

\usepackage[utf8]{inputenc}

\usepackage{microtype} 
\usepackage[T1]{fontenc}  
\usepackage{hyperref}    
\usepackage{url}      
\usepackage{booktabs}    
\usepackage{amsfonts}    
\usepackage{nicefrac}    
\usepackage{multirow}
\usepackage{arydshln}

\newcommand{\ie}{\textit{i}.\textit{e}.}

\title{Reweighting Strategy based on Synthetic Data Identification 

for Sentence Similarity}

\author{Taehee Kim$^{\dagger}$\thanks{\hspace{0.2cm} Equal contribution}, ChaeHun Park\footnotemark[1], Jimin Hong,  \\ \textbf{Radhika Dua, Edward Choi} and \textbf{Jaegul Choo} \\
KAIST AI, Letsur$^{\dagger}$ \\
    \hspace{0cm}\texttt{\{taeheekim,ddehun,jimmyh,radhikadua,edwardchoi,jchoo\}@kaist.ac.kr} \\
 }

\begin{document}

\maketitle

\begin{abstract}

Semantically meaningful sentence embeddings are important for numerous tasks in natural language processing. 
To obtain such embeddings, recent studies explored the idea of utilizing synthetically generated data from pretrained language models~(PLMs) as a training corpus.
However, PLMs often generate sentences much different from the ones written by human.
We hypothesize that treating all these synthetic examples equally for training deep neural networks can have an adverse effect on learning semantically meaningful embeddings.
To analyze this, we first train a classifier that identifies machine-written sentences, and observe that the linguistic features of the sentences identified as written by a machine are significantly different from those of human-written sentences.
Based on this, we propose a novel approach that first trains the classifier to measure the importance of each sentence.
The distilled information from the classifier is then used to train a reliable sentence embedding model.
Through extensive evaluation on four real-world datasets, we demonstrate that our model trained on synthetic data generalizes well and outperforms the existing baselines.\footnote{Our implementation is publicly available at \url{https://github.com/ddehun/coling2022_reweighting_sts}.}

\end{abstract}

\begin{table*}[t]
\footnotesize
\centering
    \begin{adjustbox}{width=0.85\textwidth}
        \begin{tabular}{cccccccccc}
            \toprule
            &\multicolumn{3}{c}{STSb} & \multicolumn{3}{c}{QQP} & \multicolumn{3}{c}{MRPC} \\ 
            \cmidrule(lr){2-4} \cmidrule(lr){5-7} \cmidrule(lr){8-10}
            & $x_{h}$ & $p_{D}(x_m)$ $\uparrow$ & $p_D(x_m)$ $\downarrow$ & $x_h$ & $p_D(x_m)$ $\uparrow$ & $p_D(x_m)$ $\downarrow$ & $x_h$ & $p_D(x_m)$ $\uparrow$ & $p_D(x_m)$ $\downarrow$ \\ \midrule
            BLEU-N       & 34.80 & 25.75 &	\underline{2.93} &	30.3  &	34.95&	\underline{7.86} &	48.53&	46.97&	\underline{5.59} \\ 
            Jaccard      & 41.98 & 33.97 &	\underline{5.98} &	39.91 &	42.49&	\underline{11.31}&	53.55&	53.33&	\underline{10.52} \\
            Distinct-N   & 44.53 & 35.93 &	\underline{17.03}&	38.10 &	25.23&	\underline{24.10}&	44.63&	32.10&	\underline{22.00} \\ 
            Zipf coeff.  & 1.03  & 1.07 &	\underline{1.23} &	1.11  &	\underline{1.06}&	1.12 &	0.98 &	1.02 &	\underline{1.23}  \\ \bottomrule
        \end{tabular}
    \end{adjustbox}
    \vspace{-2mm}
    \caption{Results for comparing the sentences in different group. Jaccard indicates Jaccard similarity score. The score of generated sentences far from human scores is highlighted in \underline{underline}. BLEU-N and Distinct-N indicate the average score with different $N$. The full results are available in Appendix~\ref{sec:appendix:further_analysis}.}
    \label{tab:main_analysis}
\end{table*}

\section{Introduction}

High-quality sentence embeddings are essential in diverse applications of natural language processing~\citep{cer2018universal, reimers2019sentence}, including semantic textual similarity~\cite{cer-etal-2017-semeval} and paraphrase identification~\cite{dolan2005automatically}.
Unfortunately, obtaining a large amount of human-annotated datasets to train a sentence embedding model is difficult and expensive.
To address this, \citet{schick-schutze-2021-generating} recently introduced a method, DINO, to train a sentence embedding model on synthetic data generated from pretrained language models (PLMs). 
Despite the effectiveness and scalability of DINO, however, the difference between machine-written and human-written examples has not been carefully investigated.
In other words, the study on the impact of treating all these synthetic examples equally during training remains under-explored.

To this end, we first conduct an in-depth analysis to demonstrate the shift of synthetic samples from the human-written sentences. In particular, we train a classifier (\ie, Synthetic Data Identification (SDI) model) that identifies synthetic data from human-written sentences and observe that the linguistic features of the sentences predicted as machine-written are much different from the human-written sentences compared to the linguistic features of the sentences predicted as human-written.

Based on this analysis, we propose a simple method, \textbf{R}eweighting Loss based on \textbf{I}mportance of Machine-written \textbf{SE}ntence~(RISE), which first utilizes the trained SDI model to measure the importance of each sentence in learning semantically meaningful sentence embeddings for sentence similarity tasks. 
We then utilize this distilled information from the SDI model to reweight the loss of each synthetic example during training.

We extensively evaluate our method on multiple sentence similarity datasets and observe that our model outperforms all the baselines across diverse datasets, even when they are evaluated on other datasets from a distinct distribution with training datasets. Our contributions include:
\begin{itemize}
    \setlength\itemsep{0.01em}
    \item We analyze the linguistic features of machine-written sentences in synthetic dataset compared to human-written sentences.

    \item We propose a simple method that adjusts the contribution of synthetically generated samples to learn a reliable sentence encoder.

    \item We extensively evaluate our model on diverse datasets and observe that our method demonstrates consistent gains, generalizes well to datasets from different domains, and is robust to the adversarial attack.
    
\end{itemize}

\begin{table*}[t!]
\centering
\begin{adjustbox}{width=1\textwidth}
\begin{tabular}{l}
\toprule
 $x_{h}$: Mitchell, Man of La Mancha, Malcolm Gets from Amour and John Selya, Movin' Out, complete the field. \\
 $x_{m}$: The other 30 percent is going to be a challenge\textcolor{red}{,} ($y=0$) \textbf{\textit{(Grammatical Error)}} \\ 
 \hdashline
 $x_{h}$: A black dog is walking on the grass with a stick in it's mouth. \\
 $x_{m}$: \begin{tabular}[c]{@{}l@{}} A brown dog walks with his stick in his mouth and it is brown in color. The black dog is walking with his\\stick in his mouth and \textcolor{red}{it has a brown in color on it.} ($y=0.5$) \textbf{\textit{(Repetition)}} \end{tabular} \\
 \hdashline
 $x_{h}$: Janet Racicot heard the thud from the kitchen, where she was getting a glass of water, she said in an interview. \\
 $x_{m}$: I have been to a few movies and \textcolor{red}{I am not a big movie fan. I like horror movies.} ($y=0$) \textbf{\textit{(Logical Flaw)}} \\ 
\hdashline

 $x_{h}$: \begin{tabular}[c]{@{}l@{}} The rate of survival without serious brain damage is about 10 percent, said Bossaert, a professor at the\\University Hospital in Antwerp, Belgium. \end{tabular} \\
 $x_{m}$: \begin{tabular}[c]{@{}l@{}} The brain damage in the patient is not serious. The patient is a young woman who was \textcolor{red}{born in the United}\\\textcolor{red}{States, but was born in Belgium.} ($y=0.5$)  \textbf{\textit{(Logical Flaw)}} \\ \end{tabular} \\ 

\hdashline
 $x_{h}$: It would be difficult to overestimate the potential dangers of the Remote Procedure Call (RPC) vulnerability. \\
 $x_{m}$: \begin{tabular}[c]{@{}l@{}} A senior official in the Oil ministry asked an oil company's CEO about his meeting with the minister.\\\textcolor{red}{This is a very sensitive issue and is very sensitive to both sides.} ($y=0$)  \textbf{\textit{(Uncommon in Context)}} \\  \end{tabular} \\ 
 \bottomrule
\end{tabular}
\end{adjustbox}
\caption{Examples of machine-written sentences identified by the SDI model as unrealistic. The part of sentences that contains linguistic errors is highlighted in \textcolor{red}{red}. More examples are available in Appendix~\ref{sec:appendix:qualitative_analysis}.}
\label{tab:main_good_cases}
\end{table*}

\section{Related Work}
Synthetic data generation using pretrained language models has shown promising results in various natural language processing tasks~\citep{gdaug,papanikolaou2020dare,daga,edwards2021guiding,chang2021neural}.
Recently, \citet{schick-schutze-2021-generating} proposed a new method, DINO, to generate a synthetic dataset for textual semantic similarity task.
Another recent work, \citet{yoo-etal-2021-gpt3mix-leveraging} proposed a new data augmentation framework for sentence classification by leveraging a large-scale PLM~\citep{gpt3}.
However, synthetic data can be misused in malicious usage, such as fake news generation.
To prevent such a fraudulent use, recent studies~\citep{zellers2019defending,weiss2019deepfake, uchendu2020authorship, adelani2020generating} aim to detect the synthetically generated text.
On the contrary, we aims to identify unrealistic sentences from machine-written data and mitigate their influence to achieve accurate and robust learning.
While \citet{yi2021reweighting} suggested controlling weights to augmented training examples, our work mainly focuses on using only synthetic samples from PLMs.

\section{Analysis on Synthetic Sentences}
\label{sec:motivation}
This section describes the generation of the synthetic dataset, followed by training the model to identify synthetic sentences from human-written ones. 
Then, we present a novel analysis to demonstrate the shift of synthetic samples from the human-written sentences.

\begin{figure*}[t!]
    \begin{center}
        \includegraphics[width=0.75\linewidth]{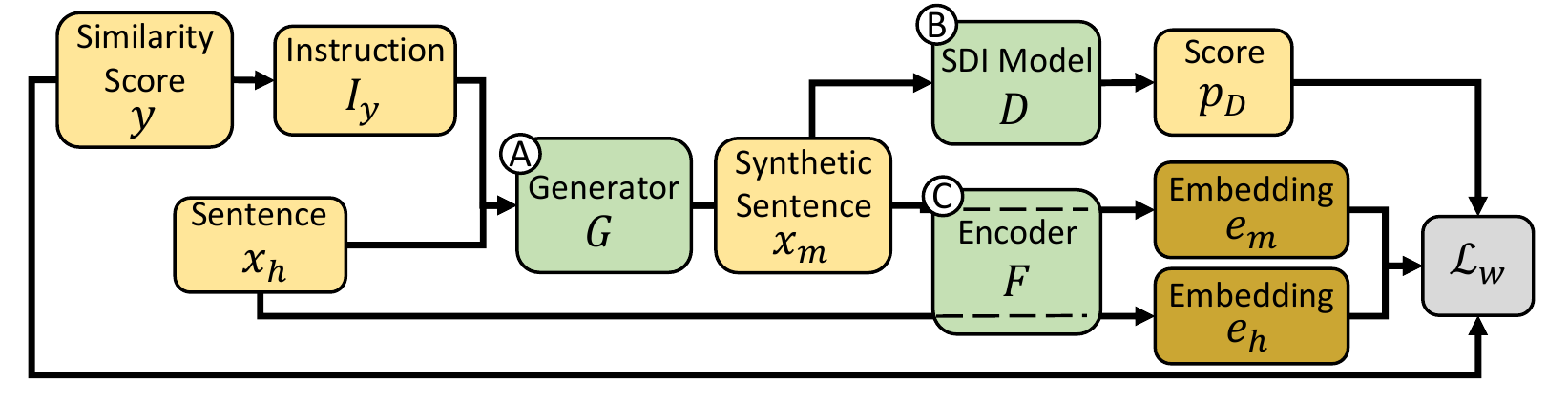}
    \end{center}
    \vspace{-7mm}
    \caption{Overview of \textbf{RISE}. We feed an instruction $I_y$ and a human-written sentence $x_h$ to the Generator $G$ which produces a machine-written sentence $x_s$. We then measure importance score $p_D$ using $x_s$ as input. Finally, we predict the similarity score using the embedding vector of $x_s$ and $x_h$. We compute the loss and multiply $p_D$.}
    \vspace{-2mm}
\label{fig:main_figure}
\end{figure*}

\noindent\paragraph{Synthetic Data Generation.}
To obtain machine-generated sentences, we leverage the ability of prompt-based zero-shot generation in a generative PLM~\citep{gpt2}~(Figure~\ref{fig:main_figure}-A). 
Specifically, given a sentence $x_h \in C_{src}$ where $C_{src}$ is a set of human-written sentences and the target similarity level $y \in Y$, this framework produces a sentence $x_m \in X_m$ that has semantic similarity with  $x_h$  equal to the target similarity level $y$.
The generated examples $\{x_h, x_m, y\}$ are later used to train a model for sentence similarity tasks. 

We use Semantic Textual Similarity benchmark~(STSb)~\cite{cer-etal-2017-semeval}, Quora Question Pairs (QQP)\footnote{\url{https://quoradata.quora.com/First-Quora-Dataset-Release-Question-Pairs}}, and Microsoft Research Paraphrase Corpus~(MRPC)~\citep{dolan2005automatically} as a source of human sentences $C_{src}$. 
We follow the details for data generation in \citet{schick-schutze-2021-generating} with their official implementation.\footnote{\url{gpt2-xl} is used as a PLM for data generation.} Finally, we obtain about 76k, 78k, and 55k examples of STSb, QQP, and MRPC datasets, respectively.

\noindent\paragraph{Synthetic Data Identification~(SDI).}
We now train a binary classification model $D$ based on a bi-directional PLM~\citep{devlin2019bert} to distinguish machine-written sentences from human-written sentences~(Figure~\ref{fig:main_figure}-B). 
We refer to this model as the Synthetic Data Identification (SDI) model and train it separately for each $C_{src}$.
We use machine-written sentences $X_m$ and human sentences $X_h$ in the same proportion for training.\footnote{The accuracy of classifiers of each dataset on the validation set are 77.87, 83.21, and 93.05\% in STSb, MRPC, and QQP datasets, respectively.} 
We use the prediction confidence $p_D$ of the generated sentence to measure how natural the sentence is.

\noindent\paragraph{Analysis.}
We now analyze to demonstrate the shift of synthetic samples from the human-written sentences. We use the following metrics to analyze the lexical-level linguistic patterns of each sentence: (1) \textbf{BLEU}~\citep{papineni2002bleu} and \textbf{Jaccard Similarity}~\cite{montahaei2019jointly} that calculate the lexical-level similarity between $x_m$ and its paired sentence. 
(2) \textbf{Distinct-N}~\cite{li2015diversity} that calculates the ratio of unique N-grams among the total number of N-grams in each group for $x_m$. 
(3) \textbf{Zipf coefficient}~\citep{holtzman2019curious} that calculates the Zipf coefficient to analyze the vocabulary usage for $x_m$.
We utilize the prediction confidence $p_D$ from the SDI model to measure the importance of generated sentences in learning meaningful sentence embeddings. We select the top 10\%~($p_D(x_m) \uparrow$) and bottom 10\%~($p_D(x_m) \downarrow$) of the machine-written sentences based on their sorted importance and analyze their linguistic features. 

Table~\ref{tab:main_analysis} demonstrates that linguistic patterns of synthetic examples vary significantly according to their importance score $p_D(x_m)$.
Furthermore, we observe that except for Zipf coefficient in QQP dataset, generated sentences with high $p_D(x_m)$ always have scores close to the scores of human-written sentences~($x_h$) compared to the sentences with low $p_D(x_m)$.\footnote{We provide a more detailed analysis in Appendix~\ref{sec:appendix:further_analysis}.}
Further qualitative analysis in Table~\ref{tab:main_good_cases} reveals that the sentences with low importance score are \textit{unrealistic} since they often contain repetition, logical flaw or expressions that a human does not use frequently.
For example, as shown in the second example of Table~\ref{tab:main_good_cases}, a person does not like movies, but in the next sentence, the machine generates a sentence that the person likes horror movies. 
In the third example, a machine generates a sentence that a woman was born in two places.

Based on these observations, we confirm that there exist a large variance in terms of how much the sentences are shifted from human sentences. 
Therefore, it is critical to handle the generated sentences carefully so that the model is not biased to the sentences that are sufficiently different from human sentences.
In the remaining of this paper, we refer to the generated sentence as \textit{unrealistic} if they contain linguistic errors or lexical patterns different from humans.
To identify such unrealistic sentences, we leverage the importance score~($p_D$) from SDI model.
We regard sentences with lower score from the model as more \textit{unrealistic}.

\section{Proposed Method}
We now introduce a simple yet effective method, \textbf{R}eweighting Loss based on \textbf{I}mportance of Machine-written \textbf{SE}ntence~(RISE), that aims to give less importance to unrealistic machine-written sentences than realistic sentences. 
Our method consists of two stages: (1) measuring the importance of the generated sentences in learning semantically meaningful embeddings using the prediction confidence $p_D$  from the SDI model (defined in Section~\ref{sec:motivation}); 2) utilizing the importance score to control the weight of the loss for each example during training so that the model does not deviate significantly from the distribution of the human text.
Other than the loss function, the training procedure is the same as standard training of a sentence embedding model based on the bi-encoder architecture~\citep{reimers2019sentence}. 
More details on training the sentence encoder are provided in Appendix~\ref{sec:bi-encoder}.

\paragraph{Reweighting Loss using Importance Score.}
We utilize the prediction confidence $p_{D}$ from the SDI model (Section~\ref{sec:motivation}) to measure the importance of generated sentences.
In particular, we modify the loss to make the realistic machine-written examples (\ie, examples with high scores) have more contributions to the loss, whereas the unrealistic machine-written examples (\ie, examples with low score) have less contribution~(Figure~\ref{fig:main_figure}-C).
The loss of each example is defined as:

\begin{table*}[t]
\centering
    \footnotesize
    \begin{adjustbox}{width=1.0\textwidth}
        \begin{tabular}{clccccccc}
            \toprule
              &  & \multicolumn{2}{c}{\textbf{STSb}} & \multicolumn{2}{c}{\textbf{QQP}} & \multicolumn{2}{c}{\textbf{MRPC}} & \multicolumn{1}{c}{\textbf{PAWS}}  \\ \cmidrule(lr){3-4} \cmidrule(lr){5-6} \cmidrule(lr){7-8} \cmidrule(lr){9-9} 
            $C_{src}$ & \textbf{Model} & \multicolumn{1}{c}{$r$} & \multicolumn{1}{c}{$\rho$}  & \multicolumn{1}{c}{Acc.} & \multicolumn{1}{c}{F1}    & \multicolumn{1}{c}{Acc.} & \multicolumn{1}{c}{F1}    & \multicolumn{1}{c}{F1}        \\ \midrule
            &GloVe          
            &47.30	&50.70	&68.51	&	63.30	&	71.53	&	80.91	&	44.16 \\ 
            &BERT-CLS           
            &17.18	&20.30	&66.38	&	61.50	&	66.03	&	79.79	&	49.32 \\ 
            &BERT       
            &47.91	&47.29	&68.70	&	64.26	&	70.38	&	80.50	&	46.05	\\
            &BERT*           
            &74.15	&76.98	&73.10	&	67.08	&	73.39	&	81.68	&	53.91	\\ 
            &RoBERTa        
            &52.36	&54.35	&67.91	&	63.67	&	72.28	&	81.20	&	44.03	\\ 
            &RoBERTa*        
            & 74.78	&77.80	&73.56	&	67.00	&	\underline{75.76}&	\underline{82.46}	&	\underline{56.48}\\ 
            &USE*            
            & 78.72	&77.08	&73.19	&	\underline{69.27}&	67.47	&	80.35	&	45.34	\\
            &InferSent*      
            &49.53	&50.86	&68.94	&	64.13	&	65.97	&	79.32	&45.01	\\ \hline
            \multirow{2}{*}{\textit{\textbf{STSb}}}
            &DINO     
            &78.45&77.71&73.14&	68.04&	70.44	&81.16&	47.30\\ 
            &RISE      
            & \underline{\textbf{79.11}}	{\scriptsize (+0.66)}&	\underline{\textbf{78.57}}	{\scriptsize (+1.46)}&	\underline{\textbf{74.47}}	{\scriptsize (1.33)}&	\textbf{69.08}	{\scriptsize (+1.04)}&	\textbf{72.84}	{\scriptsize (+2.4)}&	\textbf{82.01}	{\scriptsize (+0.85)}&	\textbf{50.24}	{\scriptsize (+2.94)} \\
            &$\llcorner$ {\scriptsize Filtering}
            & 77.73	{\scriptsize (-0.72)}&	77.45	{\scriptsize (+0.34)}&	73.06	{\scriptsize (-0.08)}&	67.94	{\scriptsize (-0.10)}&	68.96	{\scriptsize (-1.48)}&	81.35	{\scriptsize (+0.19)}&	46.72	{\scriptsize (-0.58)}\\
            &$\llcorner$ {\scriptsize Random}
            & 79.03	{\scriptsize (+0.58)}	&78.39	{\scriptsize (+1.28)}&	73.09	{\scriptsize (-0.05)}&	68.03	{\scriptsize (-0.01)}&	71.09	{\scriptsize (+0.65)}&	81.62	{\scriptsize (+0.46)} &50.17	{\scriptsize (+2.87)}\\
            \hline
            \multirow{2}{*}{\textit{\textbf{QQP}}} 
            &DINO      
            & 64.93&	65.93&	73.20&	67.72&	70.75&	80.40&44.47	\\ 
            &RISE 
            & \textbf{78.36}	{\scriptsize (+13.43)}&	\textbf{77.13}	{\scriptsize (+11.2)}&	73.35	{\scriptsize (+0.15)}&	67.76	{\scriptsize (+0.04)}&	\textbf{72.38}	{\scriptsize (+1.63)}&	\textbf{81.35}	{\scriptsize (+0.95)}&	46.28	{\scriptsize (+1.81 )}\\
            
            &$\llcorner$ {\scriptsize Filtering}
            &65.24	{\scriptsize (+0.31)}&	66.36	{\scriptsize (+0.43)}&	\textbf{73.48}	{\scriptsize (+0.28)}	&\textbf{67.95}	{\scriptsize (+0.23)}	&69.77	{\scriptsize (-0.98)}&	80.26	{\scriptsize (-0.14)}&	43.36	{\scriptsize (-1.11)}\\
            &$\llcorner$ {\scriptsize Random}
            & 73.49	{\scriptsize (+8.56)}&	72.88	{\scriptsize (+6.95)}&	73.14	{\scriptsize (-0.06)}&	67.75	{\scriptsize (+0.03)}&69.76	{\scriptsize (-0.99)}&	80.83{\scriptsize (	+0.43)}&	\textbf{46.97}	{\scriptsize (+2.5)}\\
            \hline
            
            \multirow{2}{*}{\textit{\textbf{MRPC}}}
            &DINO     
            & 75.51	&	73.87&	71.85&	65.70		&71.57	&81.55&	47.35\\
            &RISE      
            & \textbf{77.47}	{\scriptsize (+1.96)}&	\textbf{76.86}	{\scriptsize (+2.99)}&	\textbf{74.23}	{\scriptsize (+2.38)}&	\textbf{68.82}	{\scriptsize (+3.12)}	&71.97	{\scriptsize (+0.4)} &	\textbf{81.95}	{\scriptsize (+0.4)}	&\textbf{49.35}	{\scriptsize (+2.00)} \\
            &$\llcorner$ {\scriptsize Filtering}
            &76.25	{\scriptsize (+0.74)}&	74.88	{\scriptsize (+1.01)}&	71.05	{\scriptsize (-0.80)}&	64.82	{\scriptsize (-0.88)}&	71.34	{\scriptsize (-0.23)}&	80.76	{\scriptsize (-0.79)}&	47.84	{\scriptsize (+0.49)}\\
            &$\llcorner$ {\scriptsize Random}
            & 76.06	{\scriptsize (+0.55)}&	74.51	{\scriptsize (+0.64)}&	72.52	{\scriptsize (+0.67)}&	66.45	{\scriptsize (+0.75)}&	\textbf{72.19}	{\scriptsize (+0.62)}&	81.71	{\scriptsize (+0.16)}&	47.56	{\scriptsize (+0.21)}\\
            
            \bottomrule
        \end{tabular}
    \end{adjustbox}
    \caption{Evaluation results of different sentence embedding models on four sentence similarity task dataset. 
    The models trained with human-annotated dataset (e.g., NLI) are marked with *. BERT and RoBERTa indicate sentence-BERT and sentence-RoBERTa, respectively. 
    We highlight the best result in each pair of $C_{src}$/evaluation datasets and the best result in overall result in each metric as \textbf{bold} and \underline{underline}, respectively. 
    The number in right bracket indicates the performance difference with DINO. For regression task, we use Pearson correlation~($r$) and Spearman's rank correlation coefficient~($\rho$) metrics are used for evaluation. Each score represents the average of five trials.}
    \label{tab:appendix_table}
\end{table*}

\vspace{-2mm}
\begin{equation}
\mathrm{\mathcal{L}_{w}}(\theta_\textit{f})= p_{D}*\mathrm{\mathcal{L}(\theta_\textit{f}),}
\label{eq:weight_loss}
\end{equation}
\vspace{-2mm}

where $\mathrm{\mathcal{L}(\theta_\textit{f})}$ denotes the original loss of the sentence encoder $F$ for a sentence similarity task, and $\mathcal{L}_{w}(\theta_\textit{f})$ denotes the modified loss by RISE.
$\theta_\textit{f}$ denotes the parameters of the sentence encoder. 
This re-weighting procedure aims to adjust the influence of training examples based on the degree of shift of the sentence from the human-written sentences.

\section{Experimental Settings}
We evaluate each model on STSb, QQP, MRPC, and Paraphrase Adversaries from Word Scrambling of Quora Question Pairs~\citep{zhang2019paws} (PAWS-QQP) datasets.
PAWS-QQP aims to evaluate the robustness of the model against adversarial attacks for the sentence similarity task. 
We provide more details on datasets and experimental setup in the Appendix~\ref{sec:appendix:dataset} and~\ref{sec:appendix:training_details}. 

We train a model to solve the sentence similarity task as a regression problem. 
However, since all datasets except for STSb only contain discrete labels, we set the threshold using the validation dataset to make a binary decision. 
We apply our method to DINO and denote it as RISE. 
In addition to experiments with RISE, we conduct experiments with the following variants: (1) \textit{Filtering}: We filter out the bottom 10\% of the machine-written sentences  based on their sorted importance. 
We then use the remaining examples for training without using our modified loss. 
(2) \textit{Random}: We randomly sample a scalar value from $U(0,1)$ for each example and use it as its importance.
DINO and the variants of our method are based on the sentence-RoBERTa-base architecture, which are fine-tuned only on synthetic datasets.
Besides, we further compare our model against the following sentence encoders that are fine-tuned on natural language inference~(NLI) dataset: Universal Sentence Encoder(USE)~\cite{cer2018universal}, InferSent~\cite{conneau2017supervised}, sentence-BERT~\citep{reimers2019sentence}, and sentence-RoBERTa. 
We also compare with the models that are not trained on human-annotated dataset, namely: GloVe~\cite{pennington2014glove}, BERT-CLS, sentence-BERT, sentence-RoBERTa.\footnote{
Results on other STS tasks by training a regressor on top of frozen embeddings are presented in Appendix~\ref{sec:appendix:additional_results}.
}

\section{Results}
Table~\ref{tab:appendix_table} report the performance of our method and the baselines on the sentence similarity task. 
We observe that our model outperforms all the other baselines including DINO that are not trained on human-annotated dataset, and sometimes even better than the models trained on human-annotated dataset (\ie, NLI).
These results support our assumption that reweighting the loss of each machine-written sentence based on its importance enhances the model's reliability and makes it less biased to unrealistic machine-written sentences.
Furthermore, we find that the improvement is usually higher when the model is evaluated on datasets from unseen domain during training.
These results imply that our method can generalize the sentence encoder trained on a synthetic dataset when evaluated on the dataset from different domains.
In addition, our model outperforms other models on the PAWS dataset, and it shows that our method makes the model robust to adversarial attacks.
In terms of the variants of our method, using the randomly sampled scalar value as an importance score usually degrades performance.
The models that filter out unrealistic examples instead of reweighting them perform worse than RISE in most cases.
Based on these observations, we confirm that training the model using RISE can enhance the reliability of the model trained on synthetic examples.

\section{Conclusions}
\label{conclusion}

In this paper, we demonstrated that the linguistic features of unrealistic machine-written sentences are different from those of human-written sentences.
Based on this observation, we proposed a novel approach to reweight the loss based on the sentence importance from synthetic data identification (SDI) model for learning semantically meaningful embeddings. 
The extensive experiments show the effectiveness and robustness of RISE compared to other baseline approaches. 

Although extensive experiments demonstrate the effectiveness of our method, adjustment of the importance of each sentence may learn an unintended bias from the classifier.
In future work, we plan to conduct an in-depth human analysis for machine-written sentences to determine if our method correlates well with human judgement or not.
Investigating the impact of unrealistic examples in other natural language applications would also be another interesting future direction.

\section*{Acknowledgements}
This work was supported by Institute of Information \& communications Technology Planning \& Evaluation (IITP) grant funded by the Korea government(MSIT) (No. 2019-0-00075, Artificial Intelligence Graduate School Program(KAIST), and No. 2020-0-00368, A NeuralSymbolic Model for Knowledge Acquisition and Inference Techniques), and the National Research Foundation of Korea (NRF) grant funded by the Korean government (MSIT) (No. NRF-2019R1A2C4070420).

\bibliography{anthology,custom}
\bibliographystyle{acl_natbib}

\appendix

\clearpage
\newpage

\appendix
\section*{Appendix}
\begin{table*}[t]
\centering
    \begin{adjustbox}{width=0.9\textwidth}

        \begin{tabular}{lccccccccc}
            \toprule
                        &\multicolumn{3}{c}{STSb} & \multicolumn{3}{c}{QQP} & \multicolumn{3}{c}{MRPC} \\ 
                                      & $x_h$ & $p_D(x_m)$ $\uparrow$ & $p_D(x_m)$ $\downarrow$ & $x_h$ & $p_D(x_m)$ $\uparrow$ & $p_D(x_m)$ $\downarrow$ & $x_h$ & $p_D(x_m)$ $\uparrow$ & $p_D(x_m)$ $\downarrow$ \\ 
                                      \midrule
            BLEU-1       & 51.02&	40.87&	\underline{7.53}&	45.94&	46.88&	13.46&	61.86&	59.17&	15.19 \\ 
            BLEU-2       & 37.55&	27.01&	\underline{2.07}&	32.25&	36.14&	7.71&	51.13&	49.36&	3.93 \\ 
            BLEU-3       & 28.51&	19.88&	\underline{1.20}&	24.19&	30.49&	5.68&	43.57&	42.42&	1.92 \\ 
            BLEU-4       & 22.10&	15.22&	\underline{0.90}&	18.80&	26.28&	4.57&	37.57&	36.92&	1.30 \\ 
            BLEU-N       & 34.80&	25.75&	\underline{2.93}&	30.3&	34.95&	\underline{7.86}&	48.53&	46.97&	\underline{5.59} \\ 
            Jaccard              & 41.98&	33.97&	\underline{5.98}&	39.91&	42.49&	\underline{11.31}&	53.55&	53.33&	\underline{10.52} \\
            Distinct-1        &8.5&	5.1&	\underline{1.8}&	5.7&	3.7&	\underline{3.4}&	7.8&	4.3&	\underline{2.5} \\ 
            Distinct-2        &49.7&	36.5&	\underline{15.0}&	39.5&	25.5&	\underline{23.4}&	48.7&	31.4&	\underline{20.1} \\ 
            Distinct-3        &75.4&	66.2&	\underline{34.3}&	69.1&	46.5&	\underline{45.5}&	77.4&	60.6&	\underline{43.4} \\ 
            Distinct-N        &44.53& 35.93&	\underline{17.03}&	38.10&	25.23&	\underline{24.10}&	44.63&	32.10&	\underline{22.00} \\ 
            Zipf coeff.      &1.03& 1.07&	\underline{1.23}&	1.11&	\underline{1.06}&	1.12&	0.98&	1.02&	\underline{1.23}  \\ \bottomrule
        \end{tabular}
        
    \end{adjustbox}
    \caption{Results for comparing the sentences in different group. Jaccard indicates Jaccard similarity score. The score of generated sentences that is far from human scores is highlighted in \underline{underline}. For BLEU-N and Distinct-N, we report the average score with different $N$.}
    \label{tab:main_analysis_appendix}
\end{table*}

\section{Detailed Analysis on Table~\ref{tab:main_analysis}}
\label{sec:appendix:further_analysis}
In this section, we present our detailed observations in Table~\ref{tab:main_analysis} and the results of` the different N-gram in BLEU and Jaccard similarity. The results are presented in Table~\ref{tab:main_analysis_appendix}.
We observe that the number of unique N-gram occurs frequently when $p_D(x_m)$ is high.
In terms of lexical similarity~(BLEU and Jaccard) with a paired sentences, the scores of synthetic sentences~$x_m$ with high $p_D(x_m)$ are higher about 20 points than those with low $p_D(x_m)$ and are similar to $x_h$. 
The distribution of word usage in generated sentences are also close to human-written sentences when predicted realistic score is high in two out of three datasets. 
Based on these observations, we confirm that even though the sentences are generated by the same machine in the same environment, there is a large variance in terms of how much the sentences are shifted from human sentences. Therefore, it is critical to handle the generated sentences carefully so that the model is not biased to the sentences that are very different from human-written sentences (\ie, unrealistic samples).

\section{Qualitative Analysis}
\label{sec:appendix:qualitative_analysis}
\begin{table*}[t!]
\centering
\begin{adjustbox}{width=1\textwidth}
\begin{tabular}{cl}
\hline
\multicolumn{2}{c}{$p_{D}(x_{m}) \downarrow$} \\ \hline
\multirow{9}{*}{\begin{tabular}[c]{@{}c@{}} Grammatical\\Error \end{tabular}}   
                         & $x_{h}$: Mitchell, Man of La Mancha, Malcolm Gets from Amour and John Selya, Movin' Out, complete the field. \\
                         & $x_{m}$: The other 30 percent is going to be a challenge, ($y=0$) \\ \cline{2-2} 
                         & $x_{h}$: A black dog is walking on the grass with a stick in it's mouth. \\
                         & $x_{m}$: \begin{tabular}[c]{@{}l@{}} A brown dog walks with his stick in his mouth and it is brown in color. The black dog is walking with his\\stick in his mouth and \textcolor{red}{it has a brown in color on it.} ($y=0.5$) \end{tabular} \\ \cline{2-2}
                         & $x_{h}$: \begin{tabular}[c]{@{}l@{}} "But the reality is that there needs to be a big structural change, " she added, "and you can't do that without\\funding." \end{tabular} \\
                         & $x_{m}$: The fact is that there needs to be a big structural change, ($y=1$) \\ \hline
\multirow{7}{*}{\begin{tabular}[c]{@{}c@{}} Logical\\Error \end{tabular}}
                         & $x_{h}$: \begin{tabular}[c]{@{}l@{}} Janet Racicot heard the thud from the kitchen, where she was getting a glass of water, she said in an interview\\Tuesday. \end{tabular} \\
                         & $x_{m}$: I have been to a few movies and \textcolor{red}{I am not a big movie fan. I like horror movies.} ($y=0$) \\ \cline{2-2}
                         & $x_{h}$: \begin{tabular}[c]{@{}l@{}} The rate of survival without serious brain damage is about 10 percent, said Bossaert, a professor at the\\University Hospital in Antwerp, Belgium. \end{tabular} \\
                         & $x_{m}$: \begin{tabular}[c]{@{}l@{}} The brain damage in the patient is not serious. \textcolor{red}{The patient is a young woman who was born in the United}\\\textcolor{red}{States, but was born in Belgium.} ($y=0.5$) \end{tabular} \\ \hline
\multirow{5}{*}{\begin{tabular}[c]{@{}c@{}}Uncommon\\in context \end{tabular}}   
                         & $x_{h}$: It would be difficult to overestimate the potential dangers of the Remote Procedure Call (RPC) vulnerability. \\
                         & $x_{m}$: \begin{tabular}[c]{@{}l@{}} A senior official in the Oil ministry asked an oil company's CEO about his meeting with the minister.\\\textcolor{red}{This is a very sensitive issue and is very sensitive to both sides.} ($y=0$) \end{tabular} \\ \cline{2-2} 
                         & $x_{h}$: Hagel defends proposed cuts in defense spending. \\
                         & $x_{m}$: \begin{tabular}[c]{@{}l@{}} \textcolor{red}{He's not going to cut the defense budget, but he is going to cut the domestic budget by more than half.} ($y=1$) \end{tabular} \\ \hline
\end{tabular}
\end{adjustbox}
\caption{Examples of machine-written sentences identified by the SDI model as unrealistic. The part of sentences that contains linguistic errors is highlighted in \textcolor{red}{red}.}
\label{tab:good_cases}
\end{table*}

\begin{table*}[t!]
\centering
\begin{adjustbox}{width=1\textwidth}
\begin{tabular}{l}
\hline
\multicolumn{1}{c}{$p_D(x_m) \downarrow$} \\ \hline
$x_h$: \begin{tabular}[c]{@{}l@{}} The Company has always made, and continues to make, exceptional customer service and customer satisfaction\\a top priority in all business practices, "AOL added. \end{tabular}                   \\
$x_m$: \begin{tabular}[c]{@{}l@{}} AOL has always been committed to providing our customers with a wide range of services and solutions that are\\innovative and useful. ($y=1$) \end{tabular}                   \\ \hline
$x_h$: The two rugged countries got 2 to 3 inches of rain between midnight and noon.                   \\
$x_m$: I was in a bad car accident last weekend. I have to go to the doctor next week to get a check up. ($y=0$)                   \\ \hline
$x_h$: They appear in the Oct. 23 issue of The New England Journal of Medicine.                   \\
$x_m$: It's not the case that all the research shows that marijuana has a positive impact on people's mental health. ($y=0$)                   \\ \hline
\end{tabular}
\end{adjustbox}
\caption{
Examples of realistic machine-written sentences on which SDI model fails to identify as realistic.}
\label{tab:failure_cases}
\end{table*}

We qualitatively analyze the sentences that the SDI model classify as unrealistic, which include the bottom 10\%~($p_D(x_m) \downarrow$) of the machine-written sentences based on their importance. In some cases, the SDI model correctly identifies them as unrealistic, and in some cases, it fails to identify them correctly as unrealistic.

As shown in Table~\ref{tab:good_cases}, the unrealistic sentences identified by the SDI model contain repetition of the same expression or are incomplete. In addition, there were cases that contain a logical defect in the sentence. For example, as shown in the fifth example of Table~\ref{tab:good_cases}, a person does not like movies, but in the next sentence, the machine generates a sentence that the person likes horror movies. In the sixth example of Table~\ref{tab:good_cases}, a machine generates a sentence that a woman was born in two places.
Furthermore, there are sentences with no grammatical or logical defects, but contain patterns that were not common in context. In the last example of Table~\ref{tab:good_cases}, the contents of the defense budget and the individual budget are generated together, and it would not be usually used in reality. On the contrary, we find some examples that the SDI model classified as unrealistic sentences, but the sentences are realistic as shown in Table~\ref{tab:failure_cases}.

\section{Experiments on other STS tasks with Frozen Embeddings}
\label{sec:appendix:additional_results}
\begin{table*}[t]
\centering
    \footnotesize
    \begin{adjustbox}{width=1.0\textwidth}
        \begin{tabular}{clcccccccc}
            \toprule
            $C_{src}$& \textbf{Model} & \textbf{STS12} & \textbf{STS13} & \textbf{STS14} & \textbf{STS15} & \textbf{STS16} & \textbf{STS-B} & \textbf{SICK-R} & \textbf{Avg.}
             \\ \midrule
            
            &GloVe*  & 52.24& 49.91 &43.36& 55.91 &47.67 &46.00 &55.02 & 50.01\\
            &BERT	& 30.88&59.90&47.73&60.28&63.73&47.29&58.22&52.58 \\
            &BERT*	&\underline{70.97}&76.53&\underline{73.19}&79.09&74.30&76.98&72.91&74.85\\
            &RoBERTa	&32.10&56.33&45.22&61.34&61.98&55.39&62.03&53.48\\
            &RoBERTa*	&70.92&73.03&70.79&78.37&73.68&77.33&\underline{74.40}&74.07 \\ 
            &USE*       &67.06&71.55&70.59&80.27&75.76&76.85&69.31& 73.05 \\
            &InferSent* &56.15 &69.57 &64.03 &74.06& 72.00 &72.06 &66.77 & 67.80\\
            \hline
\multirow{2}{*}{\textit{\textbf{STSb}}}&DINO	&69.89&79.52&70.91&79.51&\underline{\textbf{79.14}}&77.67&64.77&74.49\\
&RISE	&69.79{\scriptsize (-0.1)}&81.09{\scriptsize (+1.57)}&72.15{\scriptsize (+1.24)}&\underline{\textbf{81.04}}{\scriptsize (+1.53)}&79.05{\scriptsize (-0.09)}&78.07{\scriptsize (+0.4)}&\textbf{72.21}{\scriptsize (+7.44)}&\underline{\textbf{76.20}}{\scriptsize (+1.71)}\\
&$\llcorner$ {\scriptsize Filtering}	&67.04{\scriptsize (-2.85)}&77.03{\scriptsize (-2.49)}&69.54{\scriptsize (-1.37)}&77.81{\scriptsize (-1.70)}&76.63{\scriptsize (-2.51)}&75.99{\scriptsize (-1.68)}&65.19{\scriptsize (+0.42)}&72.75{\scriptsize (-1.74)}\\
&$\llcorner$ {\scriptsize Random}	&\textbf{70.03}{\scriptsize (+0.14)}&\underline{\textbf{81.28}}{\scriptsize (+1.76)}&\textbf{72.63}{\scriptsize (+1.72)}&79.02{\scriptsize (-0.49)}&78.87{\scriptsize (-0.27)}&\underline{\textbf{78.68}}{\scriptsize (+1.01)}&66.89{\scriptsize (+2.12)}&75.34{\scriptsize (+0.85)}\\
\hline
\multirow{2}{*}{\textit{\textbf{QQP}}}&DINO	&56.93&71.39&59.75&67.59&73.10&68.09&61.48&65.48\\
&RISE	&\textbf{59.11}{\scriptsize (+2.18)}&\textbf{78.11}{\scriptsize (+6.72)}&\textbf{70.17}{\scriptsize (+10.42)}&\textbf{77.48}{\scriptsize (+9.89)}&\textbf{78.70}{\scriptsize (+5.6)}&\textbf{77.89}{\scriptsize (+9.8)}&\textbf{71.59}{\scriptsize (+10.11)}&\textbf{73.29}{\scriptsize (+7.81)}\\
&$\llcorner$ {\scriptsize Filtering}	&58.30{\scriptsize (+1.37)}&72.32{\scriptsize (+0.93)}&62.00{\scriptsize (+2.25)}&69.76{\scriptsize (+2.17)}&73.70{\scriptsize (+0.6)}&71.36{\scriptsize (+3.27)}&62.17{\scriptsize (+0.69)}&67.09{\scriptsize (+1.61)}\\
&$\llcorner$ {\scriptsize Random}	&56.80{\scriptsize (-0.13)}&71.17{\scriptsize (-0.22)}&59.64{\scriptsize (-0.11)}&68.32{\scriptsize (+0.73)}&72.42{\scriptsize (-0.68)}&69.71{\scriptsize (+1.62)}&65.77{\scriptsize (+4.29)}&66.26{\scriptsize (+0.78)}\\
\hline
\multirow{2}{*}{\textit{\textbf{MRPC}}}&DINO	&60.74&73.11&61.38&70.95&74.85&73.61&67.70&68.91\\
&RISE	&\textbf{66.17}{\scriptsize (+5.43)}&\textbf{77.41}{\scriptsize (+4.3)}&\textbf{68.56}{\scriptsize (+7.18)}&\textbf{76.64}{\scriptsize (+5.69)}&\textbf{76.93}{\scriptsize (+2.08)}&\textbf{76.39}{\scriptsize (+2.78)}&\textbf{71.93}{\scriptsize (+4.23)}&\textbf{73.43}{\scriptsize (+4.52)}\\
&$\llcorner$ {\scriptsize Filtering}	&59.86{\scriptsize (-0.88)}&74.76{\scriptsize (+1.65)}&62.43{\scriptsize (+1.05)}&72.74{\scriptsize (+1.79)}&75.07{\scriptsize (+0.22)}&73.25{\scriptsize (-0.36)}&69.48{\scriptsize (+1.78)}&69.66{\scriptsize (+0.75)}\\
&$\llcorner$ {\scriptsize Random}	&64.36{\scriptsize (+3.62)}&76.02{\scriptsize (+2.91)}&64.62{\scriptsize (+3.24)}&73.24{\scriptsize (+2.29)}&76.01{\scriptsize (+1.16)}&75.36{\scriptsize (+1.75)}&70.78{\scriptsize (+3.08)}&71.48{\scriptsize (+2.57)}\\
            \bottomrule
        \end{tabular}
    \end{adjustbox}
    \caption{Evaluation results of frozen sentence embedding models on STS tasks. The linear regressor is trained on top of sentence embeddings from each model. The number in right bracket indicates the performance difference with DINO. We highlight the best result in each pair of $C_{src}$/evaluation datasets and the best result in overall result in each metric as \textbf{bold} and \underline{underline}, respectively. For regression tasks, we use Spearman's rank correlation coefficient ($\rho$) as an evaluation metric. }
    \label{tab:sts_other}
\end{table*}

Following previous studies~\cite{reimers2019sentence, gao2021simcse}, we evaluate the quality of each sentence embedding by using it as a feature of a classifier.
Specifically, we train a linear regressor on top of frozen sentence embeddings from each model for STS tasks.
We use SentEval~\cite{conneau2018senteval} framework on "test" setting. 
As shown in Table~\ref{tab:sts_other}, We observe that the overall trends are consistent with the previous results in Table~\ref{tab:appendix_table}. 
RISE outperforms DINO in two source corpora~(QQP and MRPC), while the results on STSb are sometimes unclear. 
Filtering out unrealistic examples performs worse than RISE in most cases. 
Finally, our model trained on STSb corpus achieves the best average score.

\section{Training Sentence Encoder for Sentence Similarity Task}
\label{sec:bi-encoder}
Sentence similarity task aims to determine the similarity between two sentences. 
It can be formulated by classifying whether the two sentences are semantically similar or not or by measuring the distance between two sentences. 
A common and scalable approach for this task is based on Bi-encoder architecture~\citep{reimers2019sentence} which involves converting the sentences into embedding vectors and then measuring the similarity between sentences by calculating the distance between them in the embedding space.

More formally, given two sentences $s_1$ and $s_2$, and their ground truth similarity score $y$, a sentence encoder \textbf{$F$} encodes the sentences, $s_1$ and $s_2$, into their embedding vectors, $e_1$ and $e_2$, respectively. 
A distance metric $d$ is then used to measure their similarity score $\hat{y}$, which is defined by:

\begin{equation}
\hat{y} = d(e_1,e_2).
\label{eq:mse_loss1}
\end{equation}

This approach aims to predict the similarity score ($\hat{y}$) close to the ground-truth similarity score ($y$) by minimizing the mean squared error~(MSE) which is given by:
\begin{equation}
\mathrm{\mathcal{L}(\theta_\textit{f})}=\sum_{i=1}^{N}(\hat{y}_i-y_i)^2,
\label{eq:mse_loss2}
\end{equation}
where $\theta_f$ is the parameter of embedding model $F$.

\section{Datasets Details}
\label{sec:appendix:dataset}
As aforementioned in Section~\ref{sec:motivation}, STSb~\citep{cer-etal-2017-semeval}, QQP, and MRPC~\citep{dolan2005automatically} are used to obtain a corpus of human-written sentences. 
The size of corpus $|C_{src}|$ is equally set to 10,000 across datasets.
The set of similarity level $Y$ is $\{0, 0.5,1\}$. We generate samples from corpus 

\begin{table}[t!]
\centering
  \begin{adjustbox}{width=0.4\textwidth}
    \begin{tabular}{c |c c c c}
    \toprule
    \textbf{Data} &\textbf{STSb} &\textbf{QQP} &\textbf{MRPC}  &\textbf{PAWS-QQP}
    \\ 
    \hline
    $X^{train}_m$ & 76.9k & 78.2k & 55.3k & -
    \\ [0.1ex] 
    \hline
    $X^{dev}_m$ & 59.2k & 78.3k & 6.3k & -
    \\ [0.1ex]
    \hline  
    $X^{dev}_{src}$ & 1.5k & 18.1k  & 0.4k & 0.3k
    \\ [0.1ex]
    \hline 
    $X^{test}_{src}$ & 1.4k & 40.4k  & 1.7k & 0.3k
    \\ [0.1ex]
    \bottomrule 
    \end{tabular}
  \end{adjustbox}
  \caption{Dataset statistics. The class distribution of MRPC, QQP, and PAWS-QQP is imbalanced.}
  \label{tab:dataset}
\end{table}

\begin{table}[h]
\centering
  \begin{adjustbox}{width=0.4\textwidth}
    \begin{tabular}{c |c c c}
    \toprule
    \textbf{Hyperparameter} &\textbf{STSb} &\textbf{QQP} &\textbf{MRPC} 
    \\ 
    \hline
    batch size & 32 & 32 & 32  
    \\ [0.1ex] 
    \hline
    learning rate & 2e-5 & 2e-5 & 2e-5
    \\ [0.1ex]
    \hline  
    number of epochs & 3 & 3  & 3
    \\ [0.1ex]
    \hline 
    temperature $\tau$ & 0.5 & 0.9  & 0.7
    \\ [0.1ex]
    \bottomrule 
    \end{tabular}
  \end{adjustbox}
  \caption{Hyperparameters used in experiments. We conduct grid search to find the best hyperparameter settings.}
  \label{tab:hyperparams}
\end{table}

\noindent\textbf{Sentence Textual Simiarlity benchmark(STSb)}
~\cite{cer2018universal} consists of sentence
pairs drawn from news, video and image captions, and natural language inference data.
Each pair is human-annotated with a continuous score from 1 to 5; the task is to predict these scores.
In this experiment, we normalize the original similarity score to have from 0 to 1.
We evaluate using Pearson and Spearman correlation coefficients.

\noindent\textbf{Quora Question Pairs(QQP)}
~\footnote{\url{https://quoradata.quora.com/First-Quora-Dataset-Release-Question-Pairs}} consists of question pairs from the community Quora.
The task is to classify that a pairs of question have semantically same meaning.

\noindent\textbf{Microsoft Research Paraphrase Corpus(MRPC)}
~\cite{dolan2005automatically} is a corpus of sentence pairs from online news sources, with human annotations for whether the sentences in the pair are semantically same. 
The class have the imbalanced distribution.(68\% positive).

\noindent\textbf{Paraphrase Adversaries from Word Scrambling of Quora Question (PAWS-QQP)~\cite{zhang2019paws}} contains human-labeled and noisily labeled pairs that feature the importance of modeling structure, context, and word order information for the problem of paraphrase identification. 
The dataset has two subsets, one based on Wikipedia and the other one based on the Quora Question Pairs (QQP) dataset. In this paper, we only use examples based on QQP. The class have the imbalanced distribution.(31.3\% positive).

\section{Training Details}
\label{sec:appendix:training_details}

\textbf{Implementation Details} 
All experiments in Table 2 in the main paper is implemented in Ubuntu 18.04.4 LTS, 3090 RTX GPU with 24GB of memory, and  AMD EPYC 7702. 
The version of libraries we experiment are 3.8 for python and 1.4.0 for pytorch.
We implemented all models with PyTorch using Sentence-Transformers\footnote{\url{https://github.com/UKPLab/sentence-transformers}} library from Ubiquitous Knowledge Processing Lab.

\noindent\textbf{Training and Evaluation.}
We train a model to solve the sentence similarity task as a regression problem.
However, since all the datasets except for STSb only contain discrete labels, we set the threshold using validation dataset to make binary decision. Training a model takes 5 minutes per epoch.

\noindent\textbf{Hyperparameter Details}
The DINO are reproduced as described in the previous works. To compute sentence simiarity score, we use cosine similarity as distance metric.
We search the best hyperparameters using grid search. 
During the prediction of SDI model, we use use the temperature scaling~($\tau$)~\citep{kumar2018trainable} is applied before softmax function.
The best hyperparameters for each dataset of 
\textbf{RISE} are described in Table~\ref{tab:hyperparams}.

\end{document}